\documentclass[letterpaper]{article} 
\usepackage{aaai25}  
\usepackage{times}  
\usepackage{helvet}  
\usepackage{courier}  
\usepackage[hyphens]{url}  
\usepackage{graphicx} 
\usepackage{natbib}  
\usepackage{caption} 
\usepackage{algorithm}
\usepackage{algorithmic}

\usepackage{newfloat}
\usepackage{listings}
\DeclareCaptionStyle{ruled}{labelfont=normalfont,labelsep=colon,strut=off} 
\floatstyle{ruled}
\newfloat{listing}{tb}{lst}{}
\floatname{listing}{Listing}
\usepackage{subcaption}
\usepackage{booktabs}
\usepackage{multirow}
\usepackage{pifont}
\usepackage{amsmath}
\usepackage{amssymb}
\usepackage{xcolor}
\usepackage{verbatim}
\usepackage{amsthm}

\title{\textit{Yuan}: Yielding Unblemished Aesthetics through a Unified Network for \\ Visual Imperfections Removal in Generated Images}
\author{
    Zhenyu Yu \textsuperscript{\rm 1}\thanks{Corresponding author: yuzhenyuyxl@foxmail.com}, 
    Chee Seng Chan \textsuperscript{\rm 1}
}
\affiliations{
    \textsuperscript{\rm 1} Faculty of Computer Science and Information Technology, \\Universiti Malaya, Kuala Lumpur, 50603, Malaysia\\

}

\usepackage{bibentry}

\begin{document}

\maketitle


\begin{abstract}
Generative AI presents transformative potential across various domains, from creative arts to scientific visualization. However, the utility of AI-generated imagery is often compromised by visual flaws, including anatomical inaccuracies, improper object placements, and misplaced textual elements. These imperfections pose significant challenges for practical applications. To overcome these limitations, we introduce {\it Yuan}, a novel framework that autonomously corrects visual imperfections in text-to-image synthesis. {\it Yuan} uniquely conditions on both the textual prompt and the segmented image, generating precise masks that identify areas in need of refinement without requiring manual intervention—a common constraint in previous methodologies. Following the automated masking process, an advanced inpainting module seamlessly integrates contextually coherent content into the identified regions, preserving the integrity and fidelity of the original image and associated text prompts. Through extensive experimentation on publicly available datasets such as ImageNet100 and Stanford Dogs, along with a custom-generated dataset, {\it Yuan} demonstrated superior performance in eliminating visual imperfections. Our approach consistently achieved higher scores in quantitative metrics, including NIQE, BRISQUE, and PI, alongside favorable qualitative evaluations. These results underscore {\it Yuan}'s potential to significantly enhance the quality and applicability of AI-generated images across diverse fields. 
\end{abstract}

%
\begin{links}
    \link{Code}{https://github.com/YuZhenyuLindy/Yuan.git}
\end{links}

\begin{figure*}
    \centering
    \includegraphics[width=0.95\linewidth]{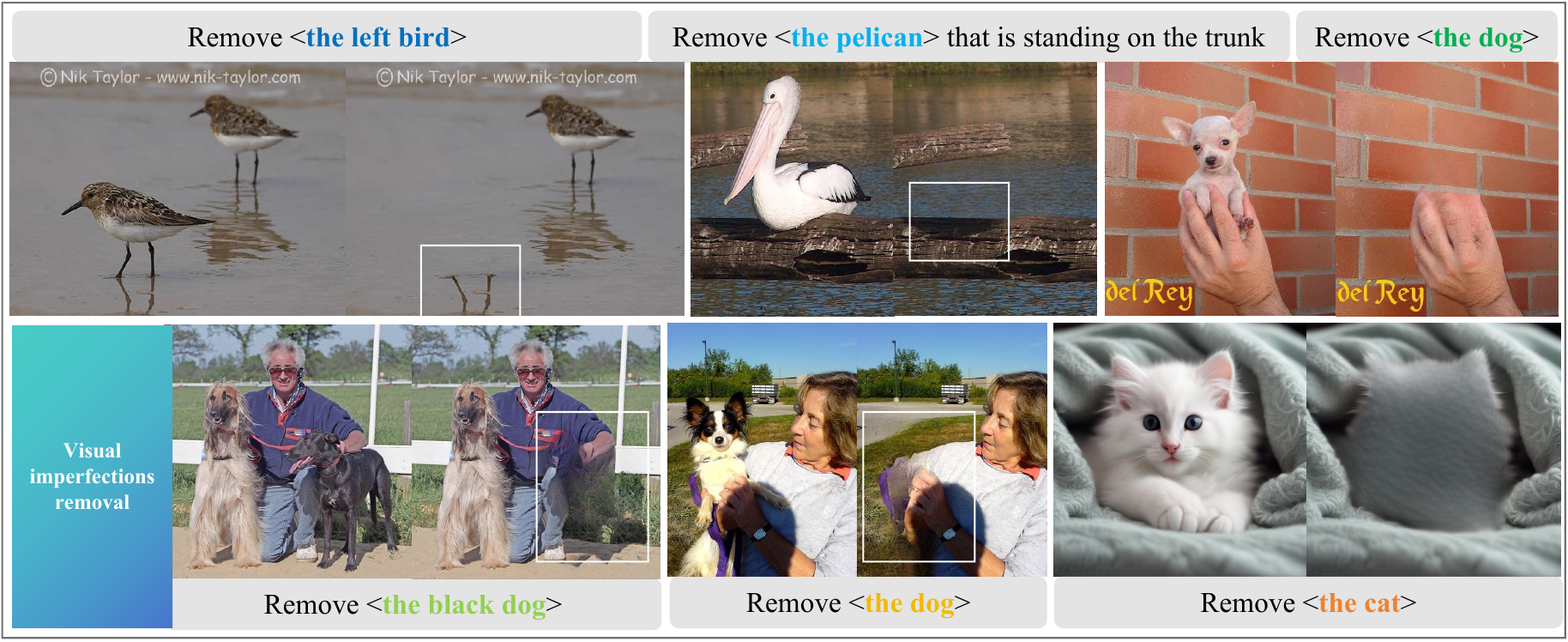}
    \caption{\textbf{Motivation for the study:} Existing algorithms for target removal often fall short in addressing related elements, such as reflections and shadows, resulting in incomplete or unnatural outcomes. Additionally, the removal of specified content can leave behind visual inconsistencies, such as unnatural postures or actions, necessitating further corrections. These challenges underscore the need for more advanced methods to achieve coherent and realistic image modifications.}
    \label{fig:why}
\end{figure*}

\section{Introduction}
The field of generative artificial intelligence (AI) has witnessed substantial advancements, especially in text-to-image synthesis, as evidenced by recent studies \cite{li2023gligen,gafni2022make,gal2022image}. These technologies enable the creation of detailed and contextually accurate images from textual descriptions\cite{yao2010i2t,cetinic2022understanding,wu2023ai}. However, they often grapple with visual imperfections such as anatomical irregularities and inappropriate textual overlays, as depicted in Fig. \ref{fig:why}. These flaws can significantly detract from the aesthetic and functional quality of the generated images.

Existing research often underestimates the importance of conditional removal of imperfections based on both text prompts and contextual image details. The ability to selectively edit or adjust elements within generated images with precision, guided by natural language, offers new opportunities in areas like interactive storytelling, bias mitigation, and ethical considerations. This approach allows users to refine visual content through linguistic cues, potentially challenging and reshaping biases within AI systems.

Current methods for addressing imperfections often rely on manual masks, which have several drawbacks: (i) they are labor-intensive and time-consuming, (ii) leading to inefficiency; (iii) their effectiveness is highly subjective and inconsistent, varying with individual skill; and they lack generalization, being limited to specific types of imperfections and not adaptable to diverse scenarios.


To address these challenges, we propose {\textit{Yuan}}, a unified framework for automatically removing visual imperfections in text-to-image synthesis outputs. {\textit{Yuan}} combines a grounded segmentation module, which identifies imperfections without predefined masks, and follow by an inpainting module that ensures contextually coherent restoration. Extensive experiments across diverse datasets demonstrate {\textit{Yuan}}'s effectiveness, validating its efficiency in tasks such as image editing and content moderation. Case studies further highlight its practical utility, offering users greater control and flexibility in image manipulation.


    
    

In summary, this paper's contributions are:

\begin{itemize}
    \item \textbf{Automated imperfection detection:} \textit{Yuan} uses a novel segmentation module to automatically detect and outline visual imperfections, eliminating the need for manual masks and improving objectivity and consistency.
    \item \textbf{Context-aware inpainting:} The inpainting module seamlessly repairs identified imperfections, preserving the visual and contextual integrity of the images and enhancing their quality.
    \item \textbf{Comprehensive validation:} \textit{Yuan} demonstrates superior performance across various datasets and scenarios, validated through quantitative metrics and qualitative assessments, proving its adaptability in diverse applications.
\end{itemize}

These contributions position \textit{Yuan} as a scalable, user-friendly solution that sets new standards for automatic visual refinement in text-to-image synthesis.

\begin{figure*}[ht]
    \centering
    \includegraphics[width=0.99\linewidth]{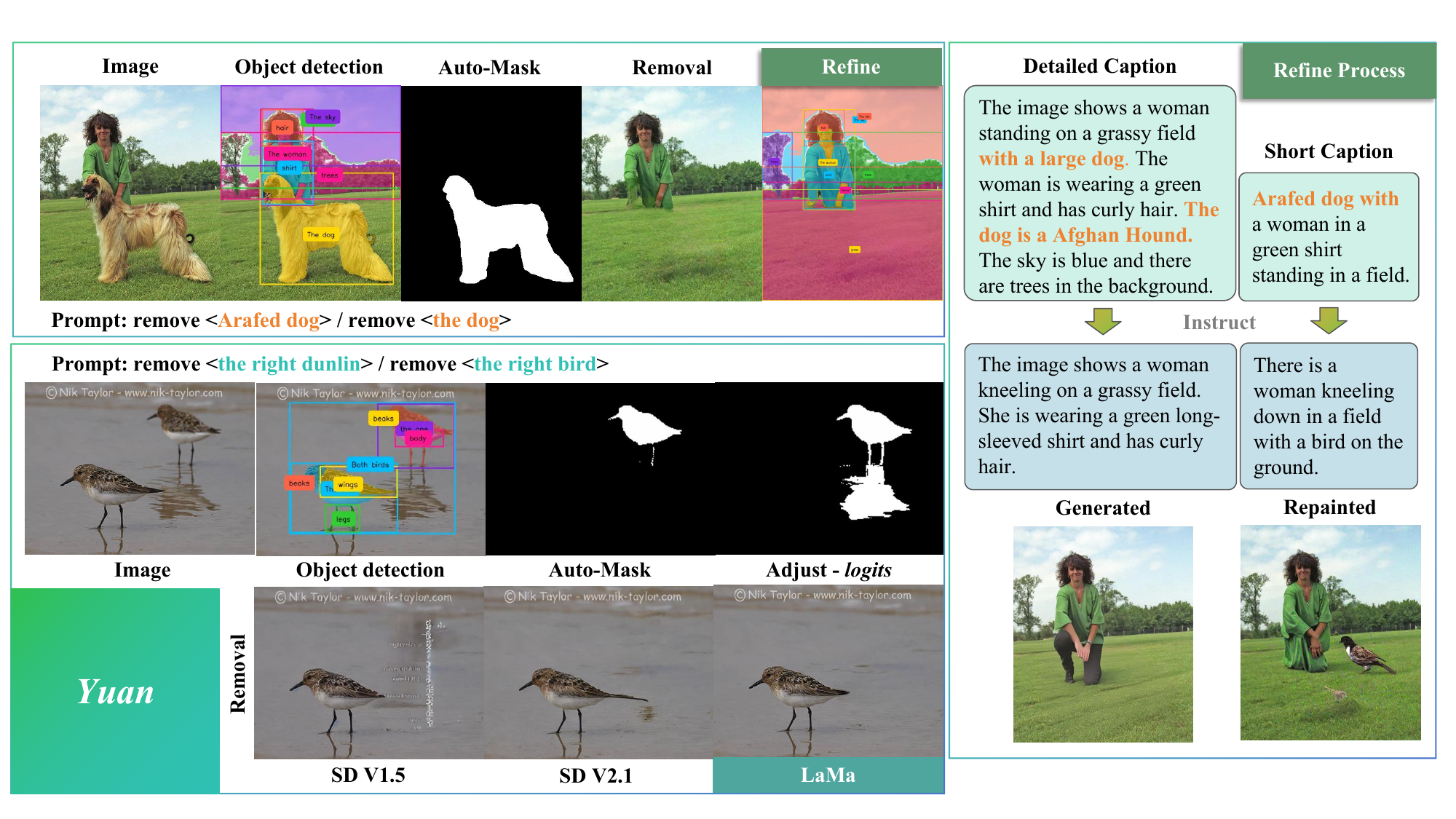}
    \caption{Our {\textit{Yuan}} framework: (a) Object detection by user prompt, (b) Automatic mask generation, (c) Object removal, (d) Inpainting and preserving original context, and (e) Refined image.}
    \label{fig:framework}
\end{figure*}

\begin{figure*}[ht]
    \centering
    \includegraphics[width=0.9\linewidth]{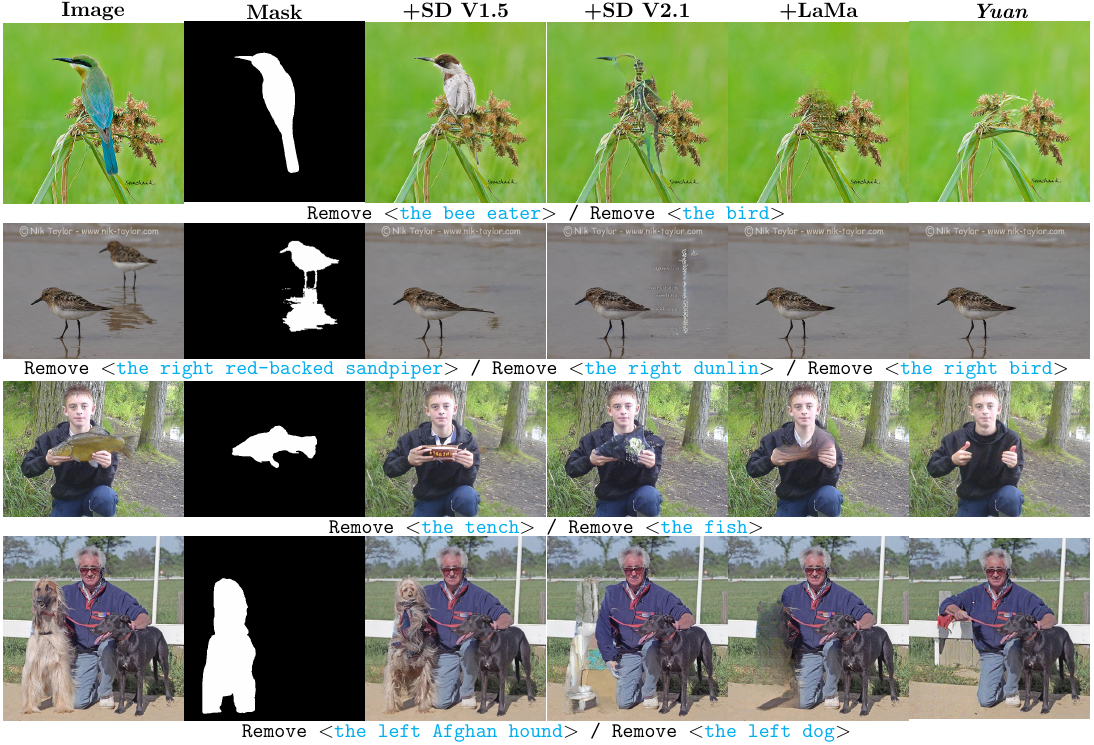}
    \caption{A comparison among Grounded SAM+SD V1.5, +SD V2.1, +LaMa, and \textit{Yuan} for different text prompt.}
    \label{fig:compare_models}
\end{figure*}

\begin{figure*}[ht]
    \centering
    \includegraphics[width=0.88\linewidth]{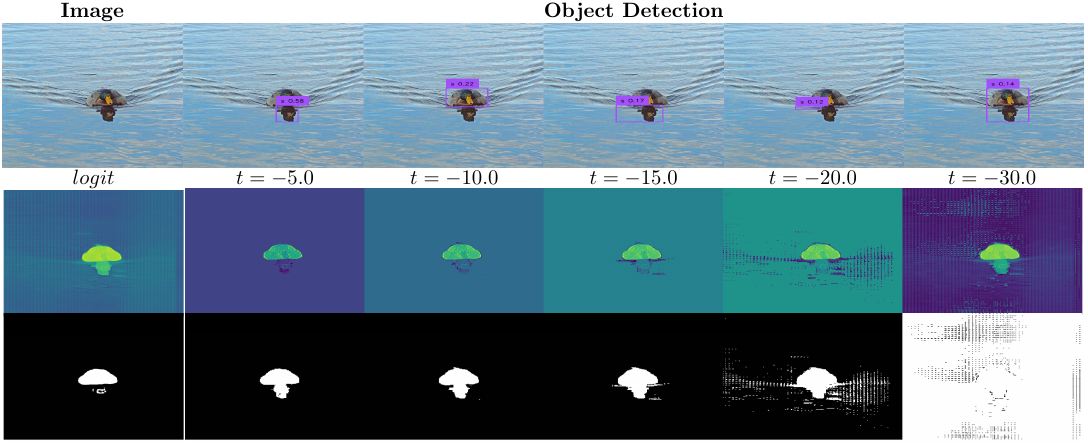}
    \caption{Ablation study results on $logits$ threshold ($t$) adjustment for automatic mask generation. }
    \label{fig:compare_logits}
\end{figure*}

\begin{figure*}[ht]
    \centering
    \includegraphics[width=0.46\linewidth]{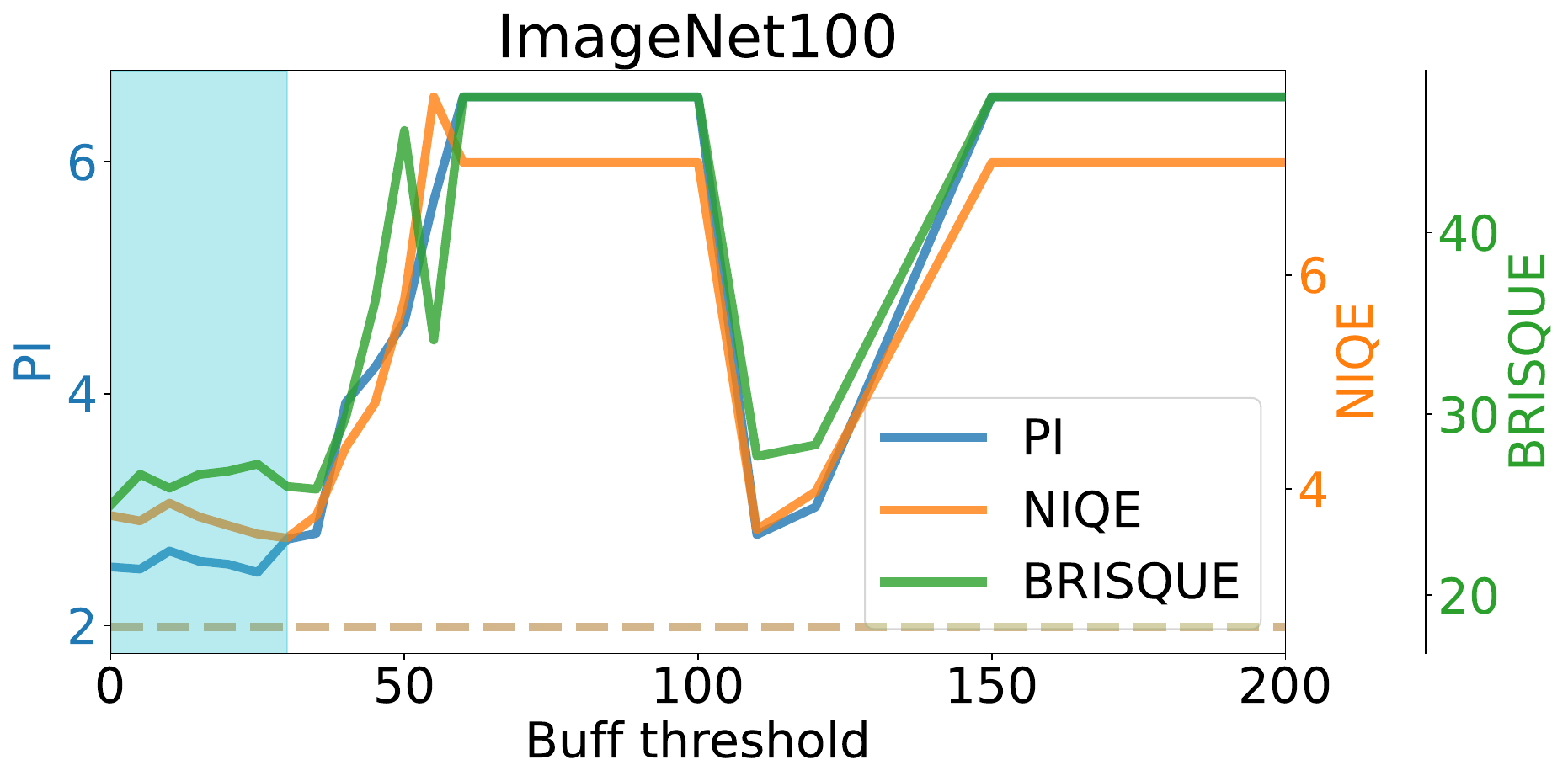}
    \includegraphics[width=0.46\linewidth]{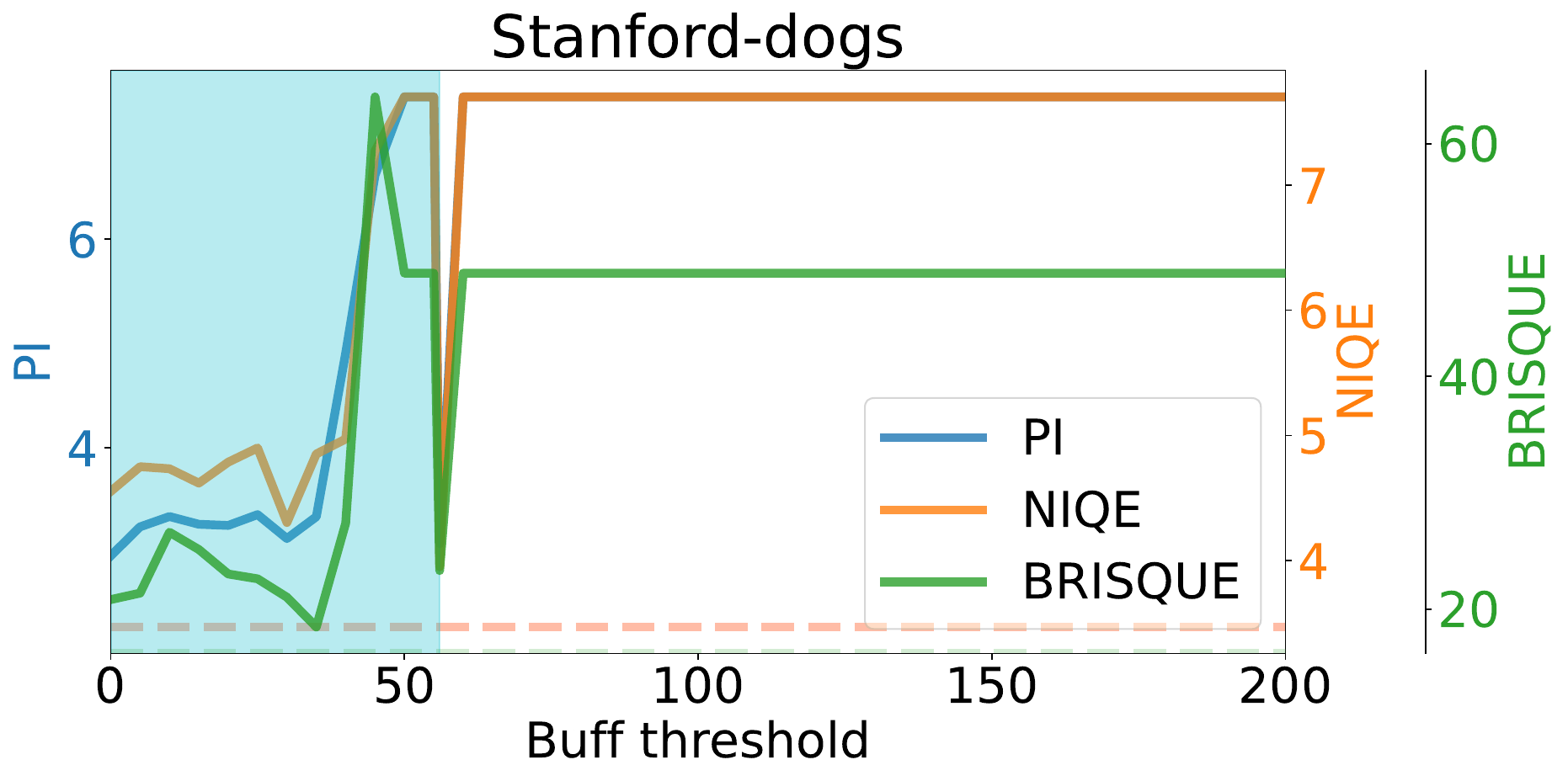}
    \caption{Ablation study results of threshold in buffer ($b$) and $logits$ ($t$). For different datasets, the threshold needs to be adjusted as needed. For \textit{Yuan}, we recommend adding two adjustable parameters, exposure $b$ and $t$, based on the original settings. This will provide convenience and service for different generated images to achieve the best results. Full figures see Fig. A3.}
    \label{fig:paint_compare}
\end{figure*}

\section{Related Work}

\subsection{Image Generation and Synthesis}
\subsubsection{Image Generation}
Generative models have revolutionized the field of image synthesis, with significant contributions from models such as Generative Adversarial Networks (GANs) \cite{goodfellow2014generative} and Variational Autoencoders (VAEs) \cite{kingma2013auto}. GANs, in particular, have achieved remarkable success in generating high-quality, realistic images through their dual-network structure, comprising a generator and a discriminator. Recent advancements, including StyleGAN \cite{karras2019style} and BigGAN \cite{brock2018large}, have further extended the resolution and diversity of generated images. Despite these successes, these models often produce outputs with visual imperfections, such as artifacts and inconsistencies, particularly in scenarios involving fine details or complex backgrounds.

\subsubsection{Text-to-image Synthesis}
Text-to-image synthesis is a rapidly advancing field focused on generating images from textual descriptions, bridging the gap between natural language and visual content. Techniques like DALL-E \cite{ramesh2021zero} and Imagen \cite{saharia2022photorealistic} use transformers and diffusion models to convert text prompts into detailed images. While these models can create intricate and contextually relevant visuals, they often face challenges such as anatomical inaccuracies and misplaced textual elements \cite{xu2018attngan}. Diffusion models \cite{ho2020denoising, song2020score}, which generate high-quality images through a reverse denoising process, have shown great promise in improving the stability and diversity of image synthesis, leading to more realistic and detailed outputs. Combining diffusion models with text encoders, as seen in DALL-E 2 \cite{ramesh2022hierarchical} and Imagen, has further enhanced the capabilities of text-to-image synthesis.

\subsection{Removal for Generated Images}

Image removal techniques focus on eliminating unwanted elements like specific objects, logos, or watermarks from digital images. Recent advancements, particularly in deep learning and generative models, have improved these processes, ensuring better image quality and realism.

\subsubsection{Concept Removal}

Concept removal is essential for privacy preservation, content moderation, and augmented reality. Recent frameworks using adversarial training and generative modeling effectively suppress sensitive information while maintaining image fidelity \cite{gandikota2024unified, wang2023security, tsai2023ring, hong2024all}. Advanced methods combine semantic segmentation with generative models to obfuscate specific objects \cite{pham2023circumventing, pham2024robust, li2024get, zhao2024separable, xiong2024editing}, and attention-based approaches enhance privacy by dynamically suppressing salient regions \cite{yang2022lightweight}.

\subsubsection{Watermark Removal}

Watermark removal is vital for repurposing images and videos legally. Traditional signal processing methods like frequency filtering often degraded image quality \cite{ray2020recent}. However, deep learning techniques, including CNNs and autoencoders, now enable more effective watermark removal and content reconstruction \cite{chen2021refit}. Despite these advances, ethical and legal concerns remain, driving the development of methods that comply with copyright laws \cite{singh2013survey}. Recent techniques focus on improving accuracy and minimizing artifacts through adversarial training and multi-scale analysis \cite{luo2023dvmark}.

Ongoing developments in concept and watermark removal are expanding the possibilities in digital image processing, with future efforts aimed at increasing robustness and adaptability across various applications.

\subsection{Image Editing and Inpainting}

Conditional image editing and inpainting have advanced significantly, enabling applications in content creation, image restoration, and augmented reality. However, challenges remain, particularly in handling shadows, complex scenes, and enhancing user interaction.

\subsubsection{Shadow Handling}

Shadows pose difficulties due to their complex interplay with light sources, objects, and backgrounds. Many algorithms struggle with light and shadow consistency, often leading to distortions when shadows are improperly removed \cite{le2019shadow}. Additionally, shadows typically have gradient edges, but current methods often produce unnatural hard edges or artifacts during editing \cite{liu2021shadow}.

\subsubsection{Editing Complex Scenes}

Editing complex scenes, especially those with multiple objects, requires algorithms to maintain spatial relationships and scene coherence. Current techniques often fail to preserve local and global consistency, resulting in edited areas that clash with the original image's color, texture, or lighting \cite{wang2020deep, zhang2021dtgan}.

\subsubsection{User Interaction}

User interaction in conditional editing tools is still limited. Systems often struggle to understand user intent, leading to results that do not align with expectations \cite{borch2022toward}. Additionally, many systems lack real-time feedback and dynamic adjustment capabilities, requiring users to make cumbersome manual adjustments \cite{sun2022ide}.

\subsection{Our Work} 

Distinct from existing methodologies, our {\textit{Yuan}} framework introduces a revolutionary automated approach to identify and correct visual imperfections by seamlessly integrating text and image data. By advancing beyond the traditional reliance on manual masking, our framework employs advanced segmentation and inpainting modules, significantly enhancing the efficiency and effectiveness of the image refinement process. This automation not only aligns with the latest developments in generative AI, but also addresses critical gaps identified in current practices, such as the labor-intensive nature of manual interventions and the inconsistencies they introduce.



\begin{table*}[ht]
    \centering
    \resizebox{0.99\textwidth}{!}{\begin{tabular}{c|cccc|cccc|cccc}
    \toprule
        \multirow{2}{*}{\textbf{Metrics}} & \multicolumn{4}{|c|}{\textbf{ImageNet100}} & \multicolumn{4}{|c|}{\textbf{Stanford-dogs}} & \multicolumn{4}{|c}{\textbf{Generated-cats}} \\ \cline{2-13}
        ~ & \textbf{Image} & \textbf{+SD} & \textbf{+LaMa} & \textbf{\textit{Yuan}} & \textbf{Image} & \textbf{+SD} & \textbf{+LaMa} & \textbf{\textit{Yuan}} & \textbf{Image} & \textbf{+SD} & \textbf{+LaMa} & \textbf{\textit{Yuan}} \\ \midrule
        NIQE$\downarrow$ & \textit{3.7425}  & 5.2829  & \underline{3.0905}  & \textbf{3.0890}  & \textit{3.3380}  & 4.6187  & \underline{4.0785}  & \textbf{3.4691}  & \textit{5.2829}  & 6.2217  & \textbf{5.0716}  & \underline{5.2465}  \\ 
        BRISQUE$\downarrow$ & \textit{26.6525}  & 32.1852  & \textbf{24.7853}  & \underline{25.6086}  & \textit{9.6086}  & 25.1237  & \underline{22.0275}  & \textbf{16.2062}  & \textit{32.1852}  & \textbf{37.4372}  & 45.6096  & \underline{39.4333}  \\ 
        PI$\downarrow$ & \textit{2.5558}  & 5.9084  & \underline{2.0921}  & \textbf{2.0124}  & \textit{2.2204}  & 3.2685  & \underline{2.6190}  & \textbf{2.2841}  & \textit{5.9084}  & 6.7089  & \underline{5.5097}  & \textbf{5.4679}  \\ \bottomrule
    \end{tabular}
    }
    \caption{Comparison of object removal performance across different models. It compares the performance of Grounded SAM+SD, +LaMa, and {\textit{Yuan}} on object removal tasks across three datasets: ImageNet100, Stanford-dogs, and Generated-cats. }
    \label{table:compare_models}
\end{table*}

\section{Proposed Method - \textit{Yuan}}

\subsection{Overview}

As illustrated in Fig. \ref{fig:framework}, given a synthetic image generated by any text-to-image (T2I) model, our proposed method, {\textit{Yuan}}, employs the models to conditionally analyze the prompt and automatically generate segmentation masks. These masks are then used to selectively preserve or modify specific regions of the image, ensuring the integrity and coherence of the original visual content. The process integrates the strengths of advanced segmentation and object detection models, followed by a robust inpainting approach to maintain the original context as illustrated in Algorithm \ref{algo1}.

\subsection{Automatic Mask Generation}

In order to create an automated process for generating masks based on the synthesis prompt, {\textit{Yuan}} utilised grounded SAM. This is because it combines the precise object detection capabilities of Grounding DINO with the powerful segmentation abilities of the Segment Anything Model (SAM). This integration removes the need for manual intervention. The process begins with Grounding DINO, which uses a transformer-based architecture to detect objects in detail. Its loss function, ($\mathcal{L}_{GDINO}$), includes components for both classification ($\mathcal{L}_{cls}$) and localization ($\mathcal{L}_{loc}$), ensuring accurate detection. Once the objects are detected, the SAM model takes over to segment the identified regions. By conditioning on both the synthesis prompt and image features, Grounded SAM will auto generate precise segmentation masks (\( M_{SAM} \)). This automated approach addresses the limitations of manual methods, enhancing consistency and precision in identifying regions of interest (details see Appendix).

\subsection{Inpainting for Image Preservation}

To preserve the original characteristics of the image, we adopt the LaMa Inpainting model over traditional diffusion-based methods. This is because diffusion-based techniques often introduce inconsistencies and artifacts that can detract from the image's coherence. In contrast, the LaMa model focuses on inpainting, which involves restoring specific masked regions based on the surrounding context. The LaMa model is optimized to inpaint large masked regions effectively, predicting and filling these areas while maintaining visual consistency. This process is governed by a loss function ($\mathcal{L}_{inpaint}$) that balances reconstruction and perceptual similarity. \( \mathcal{L}_{recon} \) ensures the inpainted region matches the original image's appearance, and \( \mathcal{L}_{perc} \) maintains high-level perceptual similarity. The parameter \( \beta \) controls the balance between these two objectives. This inpainting approach ensures that modified regions blend seamlessly with untouched areas, maintaining the original image's visual integrity and coherence (detail in Appendix).
\begin{equation}
    I_{inpaint} = \text{LaMa}(I, M_{SAM})
\end{equation}

\subsection{Refining Visual Imperfects}
Our approach to refining visual imperfections consists of two key steps: (i) adjusting the output logits of the SAM to obtain more accurate masks, and (ii) employing Prompt-to-Prompt techniques for image repainting. 




\subsubsection{Adjusting Logits for Improved Masks}

Mask generation is vital for identifying regions of interest in image processing. Initially, a threshold \( t = 0 \) was used, but it often missed out shadowed areas, leading to incomplete masks. Lowering the threshold \( t \) improved feature coverage:

\begin{equation}
    {\Delta}M_{SAM}(x) = 
        \begin{cases} 
        1, & \text{if } logit(x) \geq t \\
        0, & \text{otherwise}
        \end{cases}
\end{equation}

Experiments demonstrated that setting \( t = -10 \) typically produces the best results, although the optimal value may vary depending on image complexity. This threshold is adjustable by the user, allowing for improved mask coverage, especially in images with complex backgrounds.

\subsubsection{Repainting via Prompt Instruct}

When the adjusted masks fall short of the desired refinement, we employ Prompt-to-Prompt techniques for further optimization, guiding the model by analyzing semantic differences between the original and repainted images. To create a high-quality training dataset, we use Florence2 \cite{xiao2024florence} to generate captions that provide detailed semantic information, enabling the identification of differences between the original ($C_{I}$) and repainted ($C_{r}$) images. This data, combined with user modification requests ($P$), is used to fine-tune a GPT model to map between the original and inpainted captions, resulting in a refined caption ($\Delta C_{r}$) that serves as the prompt for T2I models such as Stable Diffusion.

\begin{equation}
    \text{Caption}(I) \rightarrow {C_{I}},\quad \text{Caption}(I_{inpaint}) \rightarrow {C_{r}}
\end{equation}
\begin{equation}
    {GPT}_{ft}(P, C_{I}) \rightarrow {\Delta C_{r}}
\end{equation}

Finally, the fine-tuned ${GPT}_{{ft}}$ interprets and executes image optimization instructions by generating captions that guide Stable Diffusion in producing refined images. Empirically, we found that this approach improves visual consistency and quality, especially in cases that require extensive content modification, ensuring that the final images align with the user's intent and aesthetic objectives.

\begin{algorithm}[t]
\caption{{\textit{Yuan}} - Object Removal}
\begin{algorithmic}
\REQUIRE {
    Synthetic image \( I \) from any T2I model\\
    Prompt \(P\) from user input
}
\ENSURE Refined image \( output \)
\STATE \( D_{GDINO} \leftarrow \text{GDINO}(I,P) \) \COMMENT{Detect objects}
\STATE \( M_{SAM} \leftarrow \text{SAM}(D_{GDINO}) \) \COMMENT{Generate mask}
\STATE \( I_{masked} \leftarrow \text{Apply } \Delta M_{SAM} \text{ to } I \)
\STATE \( I_{inpaint} \leftarrow \text{LaMa}(I_{masked}, \Delta M_{SAM}) \) \COMMENT{Inpaint}
\STATE \(output \leftarrow I_{inpaint} \)
\IF{\( I_{inpaint} \) is insufficient}
    \STATE \(\Delta M_{SAM} \leftarrow logit(t)\) \COMMENT{Adjust mask}
    \STATE \(\Delta I_{masked} \leftarrow \text{Apply } \Delta M_{SAM} \text{ to } I \)
    \STATE \( I_{inpaint2} \leftarrow \text{LaMa}(\Delta I_{masked}, \Delta M_{SAM}) \) \COMMENT{Inpaint}
    \STATE \(output \leftarrow I_{inpaint2} \)
    \IF{\( I_{inpaint2} \) is insufficient}
        \STATE \(C_I \leftarrow \text{Caption($I$)}\) \COMMENT{Generate caption}
        \STATE \(C_r \leftarrow \text{GPT}_{\text{fine-tuned}}(P, C_I) \) \COMMENT{Generate new caption}
        \STATE \( I_{refined} \leftarrow \text{Generate}(\Delta C_r, I) \)
        \STATE \(output \leftarrow I_{refined} \)
    \ENDIF
\ENDIF
\RETURN \( output \)
\end{algorithmic}
\label{algo1}
\end{algorithm}

\begin{figure*}[ht]
    \centering
    \includegraphics[width=0.99\linewidth]{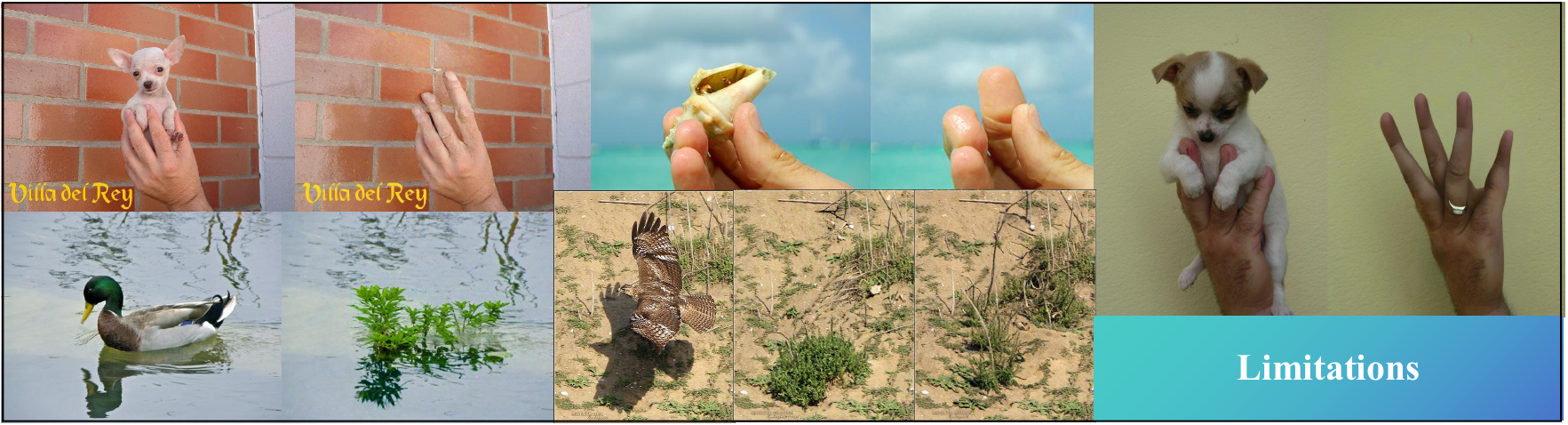}
    \caption{Limitations of {\textit{Yuan}}. The challenge of accurately rendering human hands due to complex anatomy, and the generation of unintended content during the refinement process.}
    \label{fig:compare_limitations}
\end{figure*}

\section{Experiment}
\subsection{Dataset Description}
We conducted experiments using three datasets: ImageNet100~\cite{ILSVRC15}, Stanford-dogs~\cite{khosla2011novel}, and Generated-cats. ImageNet100, a subset of ImageNet1K, includes 100 categories with 60,000 training images and 10,000 validation images, providing a condensed but representative dataset for model evaluation. Stanford-dogs contains 120 dog breeds with 20,580 images, designed for fine-grained classification. Generated-cats is a dataset created using a stable diffusion model with the prompt $<\texttt{cat}>$ (details in Appendix).

\subsection{Experimental Settings}
The experiments were conducted on a single NVIDIA GeForce RTX 4090 GPU with 24 GB of memory. For object detection, Grounding DINO used initial box and text thresholds of 0.1. In the ImageNet100 dataset, classification labels were used as text prompts, while in the Stanford-dogs dataset, either classification labels or the keyword \texttt{$<$dog$>$} were used for object removal. To evaluate generalization, 100 "cat"-related sentences were generated using ChatGPT-3.5 (details in the Appendix) and used to create a custom dataset of 100 images via Stable Diffusion, with the generation process taking about 0.5 hours. The GPT model used for fine-tuning was gpt-4o-mini.


\subsection{Evaluation Metrics}
We use three no-reference image quality assessment metrics to evaluate perceptual image quality. \textbf{NIQE} assesses image naturalness and distortion based on natural scene statistics, with lower scores indicating better quality~\cite{mittal2012making}. \textbf{BRISQUE} analyzes spatial domain features to quantify distortion, where lower scores also indicate higher quality~\cite{mittal2012making}. \textbf{PI} combines NIQE and BRISQUE scores to provide an overall quality measure, with lower scores reflecting better perceptual quality~\cite{wang2004image}. These metrics allow for comprehensive evaluation across different distortion scenarios.

\subsection{Comparison}

\subsubsection{Quantitative Analysis}

Table \ref{table:compare_models} compares three models (Grounded-SAM+SD, Grounded-SAM+LaMa, and {\textit{Yuan}}) across the ImageNet100, Stanford-dogs, and Generated-cats datasets, focusing on removal quality using NIQE, BRISQUE, and PI metrics. {\textit{Yuan}} consistently outperforms the other models, particularly on the Stanford-dogs dataset, with LaMa following and SD performing the worst. NIQE and PI scores show that LaMa and {\textit{Yuan}} produce images with naturalness and realism close to the original. In BRISQUE, LaMa and {\textit{Yuan}} surpass the original images on ImageNet100, indicating high-quality outputs with fewer distortions. However, performance declines on the other datasets, especially Generated-cats, likely due to inherent distortions. Despite these challenges, {\textit{Yuan}} remains closest to the original images, demonstrating robustness and adaptability across different datasets.

\subsubsection{Inference Time}

We compared the inference times of the models used in this study (Table \ref{table:inference_time}). Florence2 is approximately 4.56 times faster than BLIP for image caption generation. Among the text-to-image (T2I) models, SD is the fastest, but its results are suboptimal. Our \textit{Yuan} framework balances operational efficiency and system overhead, leading us to choose a combination of Florence2 and SD inpaint.

\begin{table}[t]
    \centering
    \resizebox{1.0\linewidth}{!}{
    \begin{tabular}{cccccc}
    \toprule
        \textbf{Image size} & \textbf{BLIP} & \textbf{Florence2}  & \textbf{SD T2I} & \textbf{SD I2I} & \textbf{SD inpaint} \\ \midrule
        512$\times$512 & 27.53 & 6.04 & 7.99 & 12.55 & 10.92 \\ \bottomrule
    \end{tabular}
    }
    \caption{Inference time. Unit: second/image.}
    \label{table:inference_time}
\end{table}

\subsubsection{Qualitative Analysis}

Figs. \ref{fig:compare_models} and A.2
compare the performance of SD V1.5, SD V2.1, LaMa, and {\textit{Yuan}} across various text prompts for object removal. For the prompt \texttt{Remove the <dog>}, SD models leave artifacts and incomplete blending, while {\textit{Yuan}} effectively removes the object with minimal artifacts, preserving texture. In more complex scenes, SD models struggle with context, and LaMa performs slightly better, but {\textit{Yuan}} excels in maintaining context without distortion. For the prompt \texttt{Remove the <hand>}, SD models leave visible traces, and while LaMa improves on this, {\textit{Yuan}} successfully removes the object, ensuring natural appearance. Similarly, for \texttt{Remove the <trunk>}, SD models produce artifacts, and LaMa lacks fine detail handling, but {\textit{Yuan}} achieves clean removal and preserves texture. Overall, {\textit{Yuan}} consistently outperforms the other models, accurately removing objects with minimal artifacts, demonstrating significant advancements in text-guided image editing.

\subsection{Ablation Study}
\subsubsection{Buffer Zone}

The buffer zone acts as a transitional area around the removal region, smoothing edges and reducing artifacts to improve the reconstructed image's quality and natural appearance. Table A.1
and Figs. \ref{fig:paint_compare} and A.1
show the impact of varying buffer thresholds (\( b \)) from 0 to 200 across different datasets. Optimal ranges are as follows: ImageNet100 (0$\sim$30), Stanford-dogs (0$\sim$56), and Generated-cats (50$\sim$60). Results indicate diminishing returns for \( b > 50 \), as metrics like NIQE, BRISQUE, and PI stabilize. For Generated-cats, a higher \( b \) improves clarity due to feathering effects and ambiguous boundaries. While increasing \( b \) generally enhances quality, it may also reduce original detail. The study identifies optimal thresholds to balance reconstruction quality and content originality, with \( b=15 \) set as the default in uncertain cases.

\subsubsection{Binarize Adjuster}

The logit threshold (\( t \)) is crucial for determining segmentation mask sensitivity, impacting the precision of object removal. Optimal \( t \) values are as follows: ImageNet100 ($-6\sim-5$), Stanford-dogs ($-8$), and Generated-cats ($-14\sim-7$). Figs. \ref{fig:compare_logits} and \ref{fig:paint_compare}, and Table A.2
show that the best performance on NIQE, BRISQUE, and PI metrics occurs when \(-10 \leq t \leq 0\). Below \( t = -10\), quality degrades significantly, with sharp deterioration at \( t \leq -15 \). ImageNet100 and Stanford-dogs datasets show stable metric changes, while Generated-cats exhibit more variability due to the quality of generated images. Fine-tuning \( t \) is essential for balancing image quality and reconstruction effectiveness, with \( t = -10 \) set as the default for automation. This adjuster is key to enhancing naturalness and realism while minimizing artifacts.

\section{Limitations}

Despite promising results, \textit{Yuan} has two limitations (Fig. \ref{fig:compare_limitations}): \textbf{Hand generation in generated models:} Current generated models struggle with rendering human hands accurately due to their complex anatomy and variable poses, often leading to artifacts. Enhancing hand generation fidelity remains a significant challenge, requiring improvements in both model architecture and training data. \textbf{Unintended content generation after refinement:} While effective at object removal and refinement, the process can sometimes introduce unintended elements, requiring additional refinement rounds. It can be resource-intensive and impact efficiency, highlighting the need for better controls during generation to prevent such occurrences. Addressing these limitations is essential for improving the robustness and reliability of our framework. Future work should focus on enhancing generated capabilities, particularly in generating complex anatomical features, and refining processes to better meet user expectations.

\section{Conclusion}

Text-to-image synthesis has made significant strides, but generated images often suffer from visual imperfections like anatomical inconsistencies and unwanted textual elements. Traditional correction methods relying on manual masks are time-consuming and inconsistent. This paper introduces {\textit{Yuan}}, a framework that automatically addresses these visual flaws by integrating a grounded segmentation module and an inpainting module. {\textit{Yuan}} effectively identifies and corrects image imperfections without the need for manual intervention, ensuring visual and contextual coherence. Extensive evaluations demonstrate {\textit{Yuan}}'s robustness and effectiveness, making it a valuable contribution to enhancing the quality and practicality of text-to-image synthesis.

\section{Appendix}
Appendix of this paper can be found at \texttt{https://github.com/YuZhenyuLindy/Yuan.git}

\bibstyle{aaai}
\bibliography{aaai25}

\end{document}